\documentclass[10pt,twocolumn,letterpaper]{article}

 \usepackage{cvpr}              % To produce the CAMERA-READY version
\usepackage[export]{adjustbox}
\usepackage{float}
\clearpage
\usepackage[dvipsnames]{xcolor}

\UseRawInputEncoding
\definecolor{cvprblue}{rgb}{0.21,0.49,0.74}
\usepackage[pagebackref,breaklinks,colorlinks,citecolor=cvprblue]{hyperref}
\usepackage{enumitem} % For customizing lists
\usepackage{tikz} % For drawing circles
\usepackage{hyperref} % For hyperlinks
\usepackage{algorithm}
\usepackage{algpseudocode}

\title{Hybrid Tokenization Strategy for DNA Language Model using Byte Pair Encoding and K-MER Methods}

\author{
    Ganesh Sapkota\textsuperscript{*}, 
    Md. Hasibur Rahman
     \thanks{ All authors have contributed equally. Names are listed in alphabetical order.}\\
    Missouri University of Science and Technology\\
    Rolla, MO, USA\\
    {\tt\small \{gsapkota, mrpk9\}@mst.edu}
}

\begin{document}
\maketitle
%%%%%%%%%%%%%%%%%%%include sections%%%%%%%%%%%%%%%%%%%%%%%%%%%%%%%%%%%
\begin{abstract}
% This paper introduces a hybrid tokenization strategy combining Byte Pair Encoding (BPE-600) and 6-mer methods to enhance the performance of DNA language models (DLMs). While traditional tokenization approaches like k-mers effectively capture local DNA sequence structures, they face challenges such as uneven token distribution and limited global context understanding. By integrating BPE-600, which provides adaptive subword modeling, with fixed-length k-mers, the proposed method balances local and global sequence representation. Experimental results demonstrate the effectiveness of this hybrid approach, achieving superior accuracy in next-k-mer prediction tasks compared to state-of-the-art models like NT, DNABERT2 and GROVER. The study underscores the importance of advanced tokenization strategies in genomic modeling and lays a foundation for future applications in downstream DNA sequence analysis tasks.

This paper presents a novel hybrid tokenization strategy that enhances the performance of DNA Language Models (DLMs) by combining 6-mer tokenization with Byte Pair Encoding (BPE-600). Traditional k-mer tokenization is effective at capturing local DNA sequence structures but often faces challenges, including uneven token distribution and a limited understanding of global sequence context. To address these limitations, we propose merging unique 6-mer tokens with optimally selected BPE tokens generated through 600 BPE cycles. This hybrid approach ensures a balanced and context-aware vocabulary, enabling the model to simultaneously capture both short and long patterns within DNA sequences. A foundational DLM trained on this hybrid vocabulary was evaluated using next-k-mer prediction as a fine-tuning task, demonstrating significantly improved performance. The model achieved prediction accuracies of 10.78\% for 3-mers, 10.1\% for 4-mers, and 4.12\% for 5-mers, outperforming state-of-the-art models such as NT, DNABERT2, and GROVER. These results highlight the ability of the hybrid tokenization strategy to preserve both the local sequence structure and global contextual information in DNA modeling. This work underscores the importance of advanced tokenization methods in genomic language modeling and lays a robust foundation for future applications in downstream DNA sequence analysis and biological research.
\end{abstract}    
\section{Introduction}
\label{sec:intro}
 DNA sequences in genomic data may resemble language in some ways, with elements that can be compared to grammar, syntax, and semantics, but they also have unique challenges. Unlike human language, DNA sequences don’t have clear "words" or a specific direction unless looking at biological processes like transcription or replication. This makes it challenging to apply language models to genomic data directly.

Researchers have developed DNA language models (DLMs) that use different strategies to tackle this problem. HyenaDNA \cite{Nguyen2023} proposed a model capable of processing one million tokens at a single-nucleotide level, pushing the boundary of in-context learning for genomics. Some models, like Enformer \cite{Avsec2021}, avoid the need to define "words" by using convolutional layers alongside transformer blocks for specific tasks like predicting gene expression. On the other hand, foundational models like DNABERT \cite{10.1093/bioinformatics/btab083}, and Nucleotide Transformer (NT) \cite{NT_Dalla-Torre2023.01.11.523679} and LOGO \cite{LOGO_Yang2022} take a different approach. These models are pre-trained on Bidirectional Encoder Representations from Transformers (BERT) \cite{devlin2019bertpretrainingdeepbidirectional} architecture to predict masked tokens and then fine-tuned for specific genome tasks. This process requires creating a vocabulary of "words" for DNA, with each model using its method to define these tokens. 

 DNAGPT\cite{DNAGPT} used the modified GPT architecture for tasks like DNA sequence classification and regression of guanine-cytosine content. Caduceus \cite{schiff2024caduceusbidirectionalequivariantlongrange} addressed challenges like reverse complementarity equivariance using long-range Mamba block suggested by \cite{pióro2024moemambaefficientselectivestate}, while VQDNA \cite{li2024vqdnaunleashingpowervector} adopted a convolutional encoder with a vector-quantized codebook for optimal tokenization.

Traditional methods, like k-mer tokenization, are widely used in models like NT and DNABERT, effectively preserving the sequence's local sequential structure by splitting a sequence into overlapping substrings (k-mers) of a fixed length k. However, using the fixed-length k-mers, the model essentially learns the token sequence and struggles to understand broader contexts. It often creates problems like uneven token distribution, leading to rare word issues and biased training.

To tackle these problems, DNABERT2 \cite{zhou2024dnabert2efficientfoundationmodel} introduced a more flexible method called Byte Pair Encoding (BPE) \cite{bostrom2020bytepairencodingsuboptimal} that uses variable-length tokenization by breaking down the sequence into individual characters and iteratively merging the most frequent adjacent character pairs or subwords into larger units to capture DNA’s complexity better. Also, by splitting rare words into subwords, BPE generalizes well across the dataset and helps the model to understand the global context by merging meanings from widely distributed patterns.

Building on BPE-600 tokenization, models like GROVER \cite{Sanabria2024} obtain the optimal vocabulary evaluating the accuracy of 2-6 nucleotides long next-k-mers prediction as a fine-tuning task for the foundation model training. However, challenges like uneven token usage, prediction accuracy gaps, and the risk of losing important DNA tokens during tokenization remain unsolved.

To address these issues, we have proposed combining the strengths of both k-mer and BPE tokenization methods to ensure a more balanced and context-aware representation by merging k-mer and BPE-600 tokens for a given DNA sequence. It captures short and long patterns, providing a better understanding of DNA sequences' preservation of local sequence structure and the global context. 
The major contributions of this research work are highlighted below:
\begin{enumerate}
    \item We proposed a hybrid tokenization strategy that combines the token from 6-mer tokenization and the BPE method. Then, we formulated a novel way to merge unique 6-mer tokens and optimal BPE tokens obtained from 600 BPE cycles to ensure a balanced vocabulary representing the tokens from both tokenization methods.
    \item We trained a Foundation DLM model using the optimal vocabulary obtained from the hybrid tokenization approach.
    \item We evaluate the performance of the foundation model using next-kmer prediction as a fine-tuning task and obtained the accuracy of 10.78 \% for 3-mer, 10.1\% for 4-mer, and 4.12\% for 5-mer, respectively, which demonstrate the improved performance compared to the SOTA Foundational DLMs.
     
\end{enumerate}

%-------------------------------------------------------------------------

This paper is organized into five Sections. Section \ref{sec:intro} introduces the research problem and motivation and then highlights our contributions to solving the problem. Section \ref{related_works} discusses background and related works. Section \ref{methodology} describes our proposed methodology and provides a detailed system overview. Section \ref{experiments} discusses the experimental setup, dataset, evaluation metrics, performance results, and comparison of different approaches. Section \ref{limitations} discussed the limitations and challenges faced in this research work. Section \ref{conclusion} concludes the paper by highlighting the contribution and findings. Section \ref{future_works} discusses the application of proposed DLMs in downstream tasks and the future direction of research.\\

%-------------------------------------------------------------------------

\section{Background and Related Works}
\label{related_works}
\subsection{Tokenization Methods}
% \subsubsection{Single Nucleotide Tokenization}
HyenaDNA \cite{Nguyen2023} uses a tokenization method that directly processes DNA sequences at the single nucleotide level, avoiding the need for frequency-based aggregation tokenizers. This approach provides high-resolution analysis for both short and long sequences, making it especially effective for detecting single nucleotide polymorphisms (SNPs) and understanding long-range dependencies in gene expression. The model's token set includes the standard DNA bases (A, G, C, T, and N for unknown bases) along with special tokens for padding, separation, and unknown characters, which are mapped into an embedding space for further analysis.

% \subsubsection{k-mer tokenization}
\begin{figure}[!ht]
\includegraphics[width=0.45\textwidth]{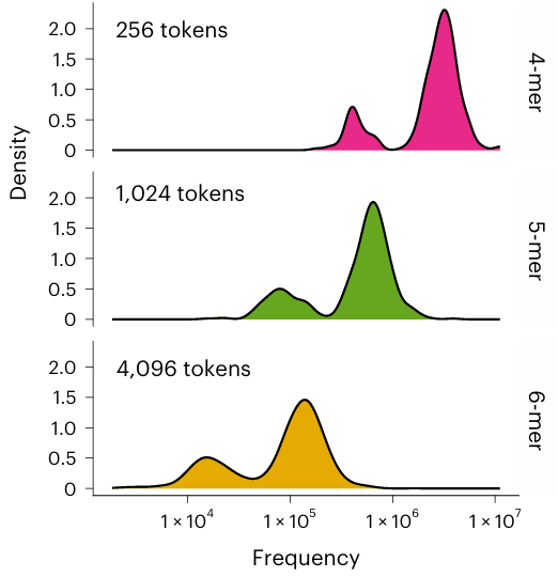}
\centering
\caption{ Showing imbalanced token distribution as a result of 
 k-mer-tokenization} 
\label{token_distribution}
\end{figure}

\begin{figure}[!ht]
\includegraphics[width=0.5\textwidth]{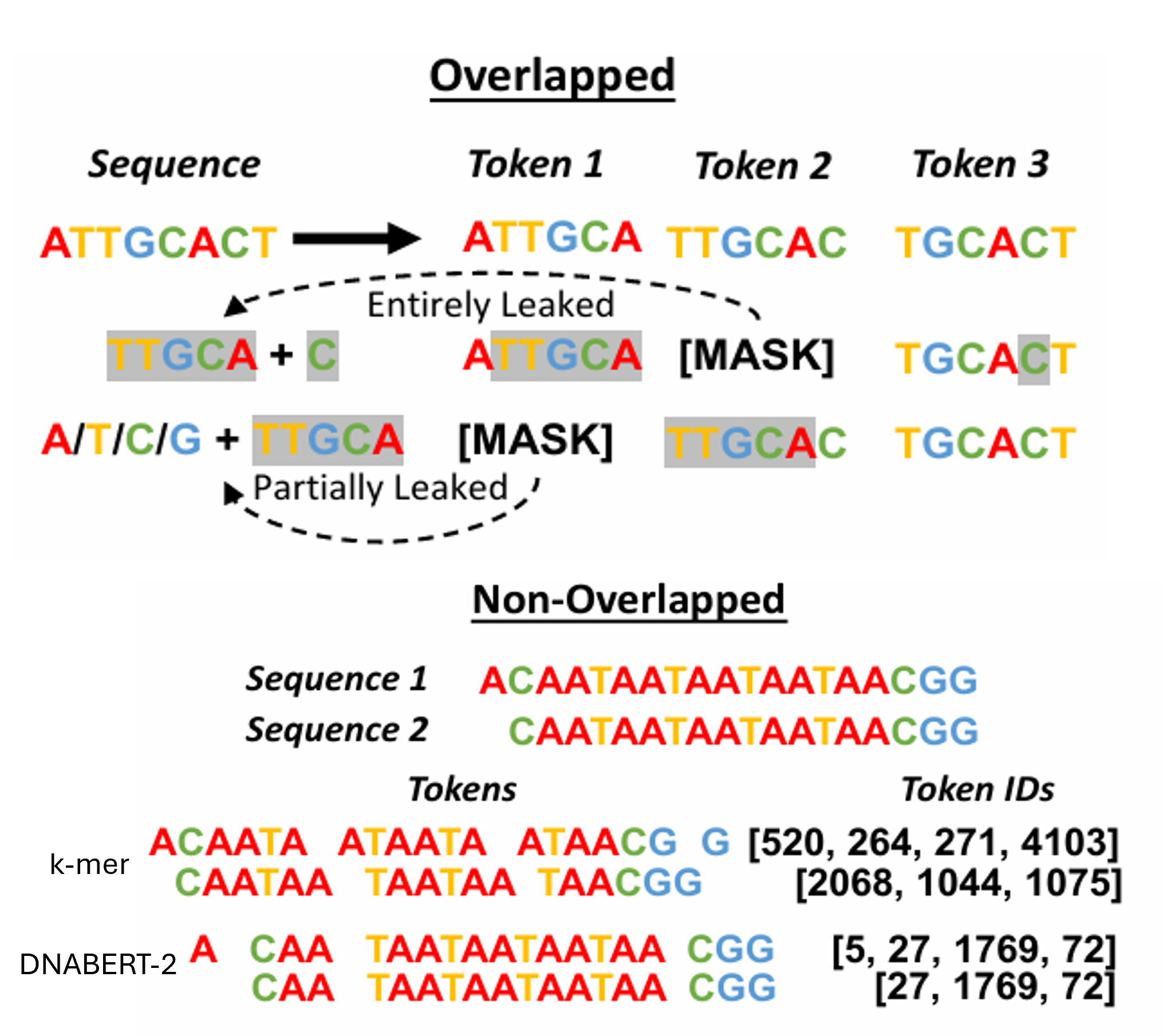}
\centering
\caption{Showing the drawbacks of overlapping and non-overlapping k-mer tokenizations. In the case of overlapping k-mer, information about a masked token is leaked by its adjacent tokens while in the non-overlapping setting, adding/deleting one nucleotide base leads to a dramatic change in the tokenized sequence.} 
\label{kmer_limitation}
\end{figure}

DNABERT \cite{10.1093/bioinformatics/btab083} employs k-mer tokenization, where DNA sequences are split into overlapping substrings of length k, called k-mers, to capture richer contextual information. For example, the sequence "ATGGCT" can be tokenized into 3-mers as ["ATG," "TGG," "GGC," "GCT"] or into 5-mers as ["ATGGC," "TGGCT"]. DNABERT trains separate models for different values of k (3, 4, 5, and 6), with each model having a vocabulary that includes all possible k-mer permutations along with five special tokens: [CLS], [PAD], [UNK], [SEP], and [MASK]. K-mers can be applied as overlapping and non-overlapping types. Overlapping K-mers aim to capture more context by including shared portions between adjacent tokens, but they often fail to provide significant advantages over single-nucleotide methods and risk information leakage. On the other hand, non-overlapping K-mers reduce the length of tokenized sequences, making them computationally efficient. Still, they can break meaningful DNA patterns into separate K-mers, potentially losing important contextual information as shown in \ref{kmer_limitation} depicted in \ref{token_distribution}.
While this method enhances context understanding at the nucleotide level, the overlapping k-mers lead to a large vocabulary with imbalanced token distribution as shown in Fig \ref{token_distribution}, and the model struggles to capture broader sequence context, focusing primarily on token-level patterns rather than the overall genomic structure. Also, the model faces a rare word problem due to the uneven distribution of the tokens. 

% \subsubsection{Byte Pair Encoding (BPE) Tokenization}
DNABERT-2 \cite{zhou2024dnabert2efficientfoundationmodel} replaces traditional k-mer tokenization with Byte Pair Encoding (BPE) \cite{bostrom2020bytepairencodingsuboptimal}, a widely used data compression and subword tokenization algorithm commonly used in natural language processing (NLP). This approach effectively addresses the limitations of k-mer tokenization, such as inefficiencies and limited contextual understanding, while preserving the computational efficiency of non-overlapping tokenization, making it a robust solution for genomic language modeling.
Fig. \ref{kmer_limitation} shows the drawback of overlapping and non-overlapping k-mer methods.

GROVER \cite{Sanabria2024} utilized BPE to obtain optimal vocabulary through the accuracy of next k-mer prediction as a fine-tuning task. Using BPE to create a frequency-balanced vocabulary, GROVER captures both token-level features and larger sequence contexts. Byte Pair Encoding (BPE) addresses the rare word problem by breaking rare or unknown words into smaller, meaningful sub-word units rather than treating them as single, unrecognizable tokens.

Built on a BERT-based transformer architecture, it is trained through masked token prediction and optimized using intrinsic validation with next-k-mer prediction, ensuring unbiased performance. GROVER excels in tasks like promoter identification, protein–DNA binding prediction, and next-k-mer prediction, outperforming existing models while providing insights into genome annotation, sequence directionality, and replication timing. However, it faces problems like uneven token usage, prediction accuracy gaps, and the risk of losing necessary DNA tokens during tokenization. Additionally, while GROVER reduces frequency imbalances, some residual effects remain, and it may prioritize token frequencies over deeper biological context in certain tasks.

% \subsection{Unigram Tokenization}

\subsection{DNA Foundation Models}
Recent advancements in DNA modeling have utilized foundation models to unravel the complexities of genomic sequences. DNABERT \cite{10.1093/bioinformatics/btab083} introduced a BERT-based approach to analyze nucleotide relationships using attention mechanisms, while the Nucleotide Transformer expanded on this by training massive models (up to 2.5 billion parameters) on data from the 1000G genomes and 850 species. DNABERT-2 \cite{zhou2024dnabert2efficientfoundationmodel} refined genome modeling by incorporating a more efficient Byte Pair Encoding (BPE) tokenizer, reducing memory and computational demands while improving performance. HyenaDNA \cite{Nguyen2023} pushed boundaries with a model capable of processing 1 million tokens at single-nucleotide resolution, pioneering in-context learning for genomics.  Caduceus \cite{schiff2024caduceusbidirectionalequivariantlongrange} addressed challenges like reverse complementarity equivariance with its long-range Mamba block, while VQDNA \cite{li2024vqdnaunleashingpowervector} adopted a convolutional encoder with a vector-quantized codebook to enhance tokenization. DNAGPT \cite{DNAGPT} extended traditional GPT architecture for tasks like DNA sequence classification and regression of guanine-cytosine content. 
Each of these models introduces unique strategies for decoding genomic data. Our research focusing specifically on improving tokenization methods to contribute further advancements in the field.

\section{Methodology}
\label{methodology}
In prior studies, the BPE-600 and 6-mer tokenization strategies have demonstrated superior performance in encoding genome sequences for training DNA language models (DLMs) such as GROVER and DNABERT. GROVER, in particular, evaluated various vocabulary configurations using a next k-mer prediction task to identify optimal tokenization approaches for genomic sequences. The results indicated that training a Byte Pair Encoding (BPE) model for 600 iterations (BPE-600) \cite{Sanabria2024} yielded improved performance in predicting the next k-mer, thereby enhancing the pre-trained DLM's efficacy in downstream tasks. However, as the value of k increased, the performance of DLMs trained with BPE-600 began to degrade. 

\begin{figure*}[!ht]
    \includegraphics[width=1\textwidth]{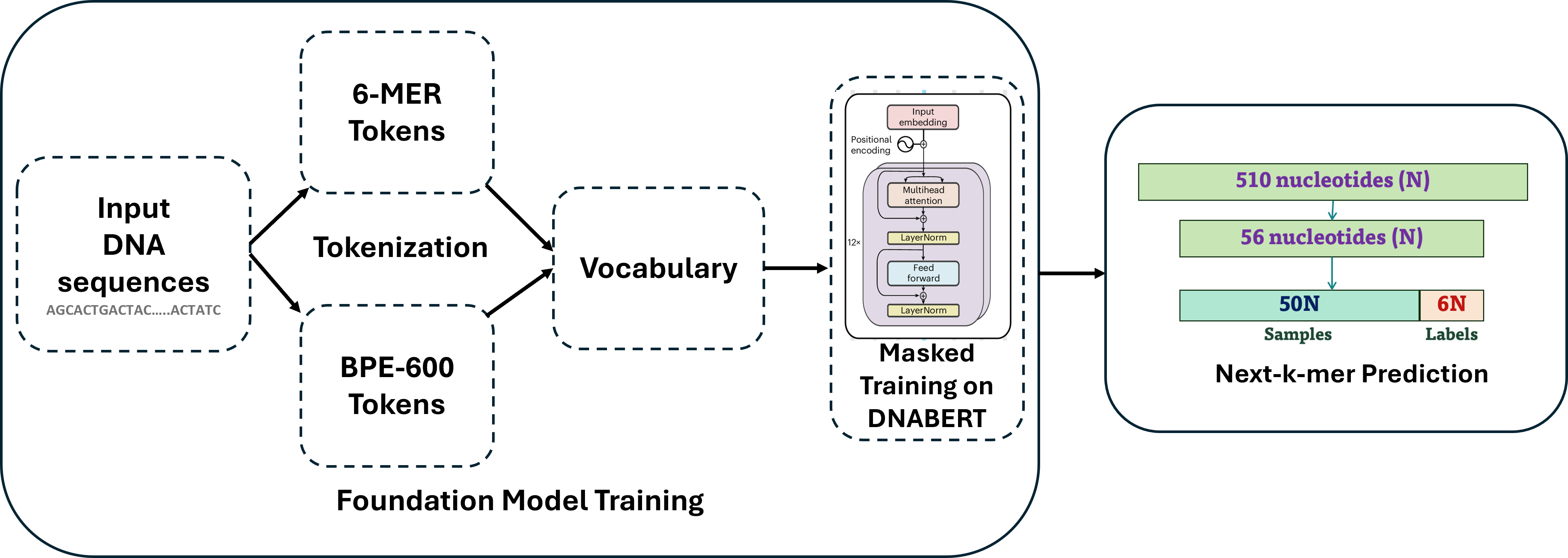}
    \centering
    \caption{An Overview of Proposed Methodology} 
    \label{project_overview}
    \end{figure*}

To address this limitation, we propose a hybrid tokenization approach combining the strengths of both 6-mer and BPE-600 tokenization methods to generate an optimal vocabulary for the DLM model.
\begin{figure}[!htbp]
\includegraphics[width=0.47\textwidth]{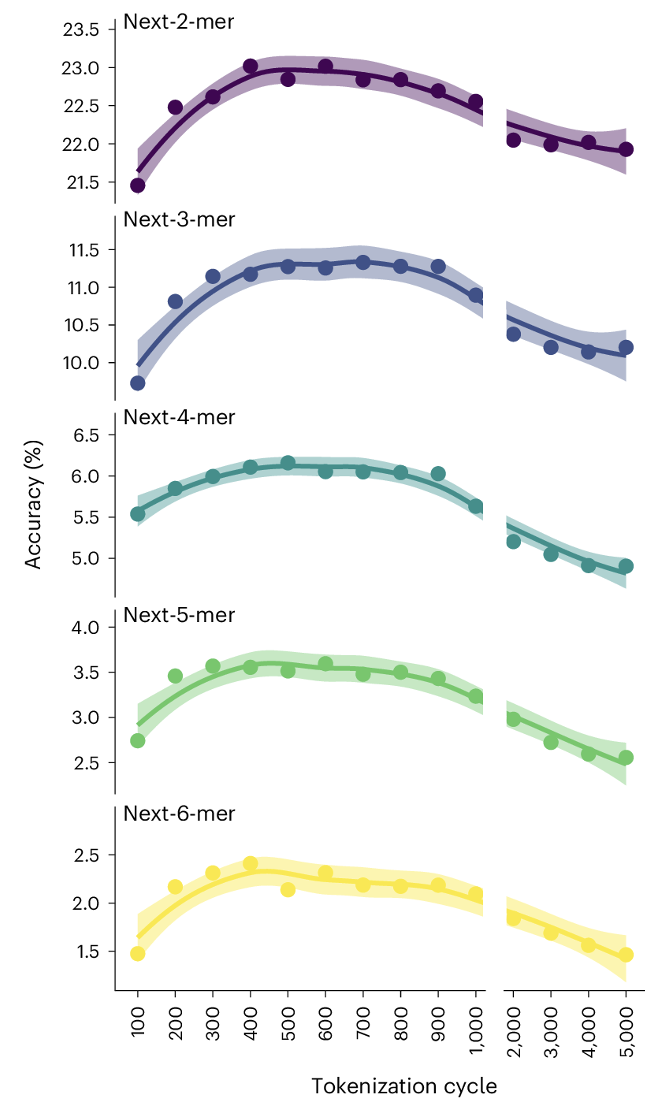}
\centering
\caption{Tokenization cycles vs k-mer prediction accuracy in BPE tokenization} 
\label{bpecycle}
\end{figure}
In our approach, the DNA sequence is first divided into segments of length 305. Each segment is tokenized using both the 6-mer and BPE-600 methods. The resulting tokens are then concatenated to form a composite representation that leverages the strengths of both tokenization techniques. This merging strategy aims to improve the next k-mer prediction performance by combining the granular, fixed-length segmentation of 6-mer with the adaptive subword modeling of BPE-600. By using k-mer prediction as the fine-tuning task, we obtain optimal vocabulary to train the DLM model. The detail overview of our methodology is depicted in Fig. \ref{project_overview}.
\begin{algorithm}[H]
\caption{Byte Pair Encoding for Genome Sequences}
\label{alg:BPE_genome}
\begin{algorithmic}[1]
\Require Set of genome sequences $D$, target vocabulary size $k$
\Procedure{BPE\_Genome}{$D, k$}
    \State $V \gets$ all unique single-nucleotide tokens in $D$ 
    \State{e.g., $\{A, T, G, C, N\}$ for DNA sequences}
    \While{$|V| < k$}
        \State $(t_L, t_R) \gets \text{Most frequent adjacent nucleotide pair in } D$
        \State{Find the most common bigram}
        \State $t_{new} \gets t_L + t_R$
        \State{Create a new token by merging $t_L$ and $t_R$}
        \State $V \gets V \cup \{t_{new}\}$
        \State{Add $t_{new}$ to the vocabulary}
        \State Replace each occurrence of $(t_L, t_R)$ in $D$ with $t_{new}$
        \State{Update sequences with the new token}
    \EndWhile
    \State \Return $V$
\EndProcedure
\end{algorithmic}
\end{algorithm}

\label{proposed method}

\subsection{BPE-600 Tokenization} 
\begin{figure}[!ht]
\includegraphics[width=0.45\textwidth]{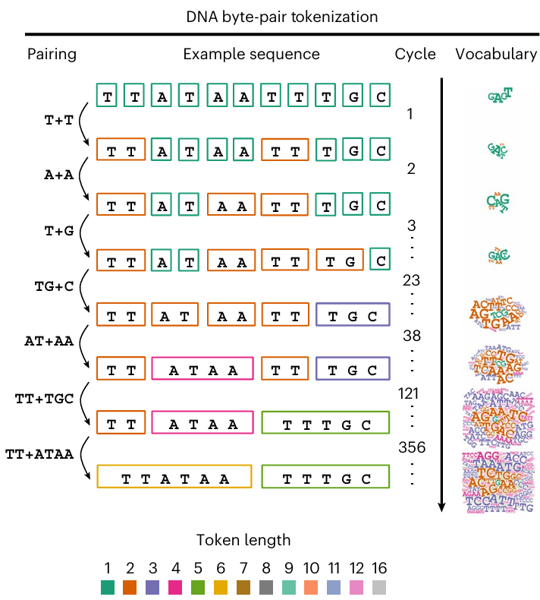}
\centering
\caption{ The BPE method highlighted 
on a sample sequence with the tokenization steps relevant to this sequence. The resulting vocabularies are colored by token length and shown in a word cloud with relative weights of the words by their frequency.} 
\label{bpe-principle}
\end{figure}
This task's Byte Pair Encoding (BPE) as shown in Algorithm \ref{alg:BPE_genome} process starts with the four basic nucleotides (A, C, G, and T), each represented as individual tokens. In the first iteration, the algorithm identifies the most frequently occurring pair of adjacent tokens within the sequence. For example, if the pair "CG" is the most common, it is merged into a new token "CG," which is then added to the vocabulary and replaces every occurrence of "CG" in the sequence. This merging process continues iteratively, with each new pair(bigram) identified and combined into a token, capturing increasingly complex patterns in the sequence as the vocabulary grows. As the BPE algorithm progresses, new tokens represent larger and more intricate substructures, allowing the model to capture higher-level dependencies within the DNA sequences. For this task, we adopted the GROVER configuration as outlined in Fig. \ref{bpe-principle} and ran the BPE process for a total of 600 iterations, which were empirically shown to yield optimal performance (see in Fig \ref{bpecycle}). After these 600 iterations, a vocabulary of 601 unique tokens was generated from the training dataset. These tokens were then consistently applied to encode the BPE-tokenized test sequences, ensuring that the representation of both training and testing data remained uniform and cohesive.
\subsection{k-mer Tokenization}
The \textit{k}-mer algorithm is designed to extract all overlapping substrings of a specified length from a list of DNA sequences. In this particular case, we focus on using a substring length of \(k = 6\), known as 6-mers. The purpose of the algorithm is to systematically process each DNA sequence by iterating through every possible starting position where a 6-mer substring can be extracted.  
Given a DNA sequence of length \(n\) (in this case, \(n = 305\)) and a specified substring length \(k = 6\), the algorithm generates overlapping substrings starting from position 0 up to position \(n - k\). Since \(n = 305\) and \(k = 6\), this means the algorithm will generate a total of \(n - k + 1 = 300\) overlapping 6-mer tokens for each segment of the DNA sequence. In other words, every DNA segment is broken into 300 distinct 6-mers, ensuring comprehensive coverage of the sequence's substrings.
To create a robust vocabulary for training purposes, we utilized the 6-mer tokenization approach on the human genome dataset, specifically selecting segments from the same dataset recommended by GROVER \cite{Sanabria2024}. This genome dataset provided a large and representative source of overlapping 6-mers across various DNA sequences, allowing us to build a comprehensive vocabulary of 6-mers. We have selected 6-mers as it shows better performance in DNABERT \cite{10.1093/bioinformatics/btab083}. These tokens were then instrumental in the subsequent modeling and tokenization processes, ensuring that our approach captured the necessary patterns and dependencies within the human genome sequences effectively.

\subsection{Hybrid Tokenization (BPE+6-mer)}
\label{hybrid-tokenization}
\begin{figure*}[!ht]
\includegraphics[width=1\textwidth]{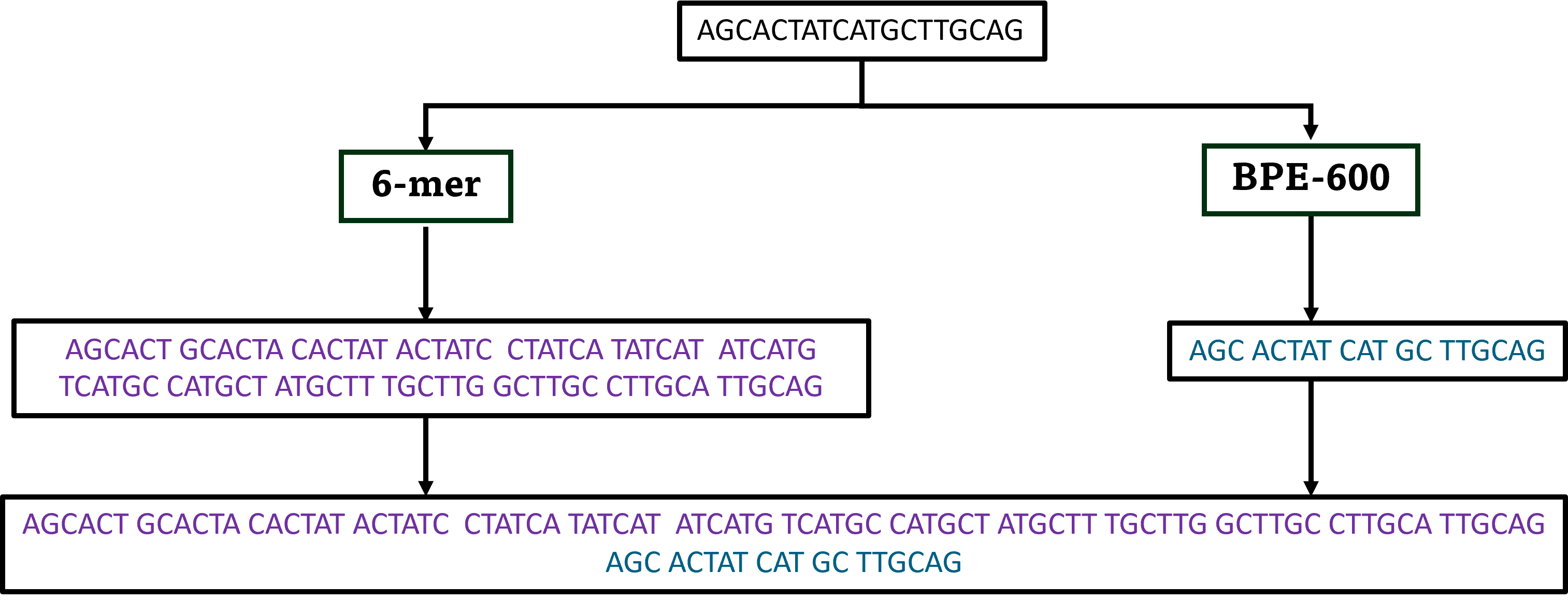}
\centering
\caption{An Overview of 6-mer BPE merging} 
\label{bpe_kmer_merging}
\end{figure*}
Our study employed two tokenization strategies—BPE and 6-mer—to preprocess DNA sequences for different modeling tasks. Initially, separate vocabulary lists were generated for each method during training. These lists were combined, and only the unique tokens were retained, resulting in a consolidated vocabulary of 4,461 unique tokens, including five special tokens. The hybrid tokenization approach uses unique tokens as follows: [CLS] marks the start of the sequence, [SEP] separates DNA sequence segments, [MASK] is used for Masked Language Modeling, [PAD] pads shorter sequences, and [UNK] represents out-of-vocabulary words. This vocabulary served as the foundation for tokenizing DNA sequences. For a DNA sequence of length n, applying BPE-600 tokenization produced a sequence of tokens with a length strictly less than n, reflecting the compression capability of the BPE algorithm. In our experiments, for a sequence of n=305, the BPE tokenized output had a maximum observed token length of 99.

In contrast, the 6-mer tokenization consistently produced a fixed output of 300 tokens for the sequence length of 305, due to its predefined segmentation logic. Upon merging the tokens generated by BPE and 6-mer tokenizations, the resulting sequence had a maximum observed token length of 399, fed into the DLM. In this work, we employed a specific masking strategy to train the foundation Deep Learning Model (DLM). For the initial 300 tokens, which correspond to k-mer tokens, we applied a localized masking approach. For each masked token, we hide the 3 tokens to the left and 2 tokens to the right, thereby preventing overlap between masked regions. This method ensured that the context of each k-mer was sufficiently occluded without introducing redundancy. We adopted a random masking scheme for the remaining tokens, where 15\% of the tokens were selected and masked randomly.

\subsection{Foundation Model Traning}
\begin{figure}[!ht]
\includegraphics[width=0.42\textwidth]{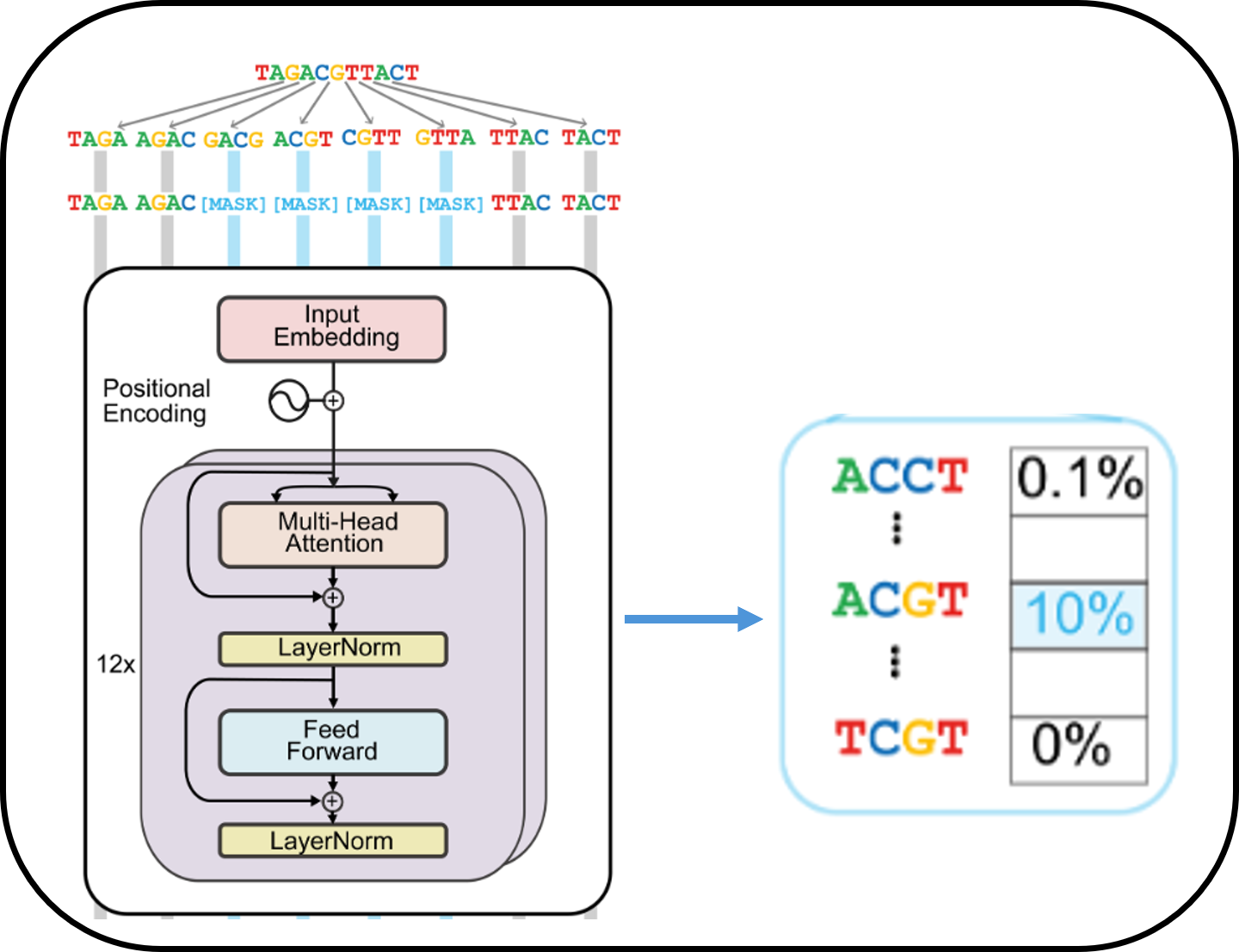}
\centering
\caption{The model architecture of DNABERT.} 
\label{dnabert_architecture}
\end{figure}
In this architecture, we extend the traditional BERT \cite{devlin2019bertpretrainingdeepbidirectional} model by incorporating an additional Long Short-Term Memory (LSTM) layer to improve sequence modeling, similar to the approach taken by GROVER \cite{Sanabria2024}. The base model is constructed using BERT with 12 Transformer layers, each consisting of 12 attention heads, enabling the model to capture rich contextual dependencies through bidirectional self-attention as depicted in fig. \ref{dnabert_architecture}. Following the BERT encoder, a single LSTM layer is introduced to process the token representations further. This LSTM layer, with 768 hidden units, allows the model to incorporate sequential dependencies, providing memory mechanisms to capture long-term contextual information that may not be fully exploited by the Transformer architecture alone. The hidden size of the LSTM layer is set to 768, aligning it with the hidden dimension of BERT, ensuring a consistent representation space across both the Transformer and LSTM components. By integrating transformers for global context and LSTM for sequential dependencies, the model balances capturing local patterns and broader relationships, making it ideal for tasks requiring context and sequence preservation.
\begin{figure*}[!ht]
\includegraphics[width=1\textwidth]{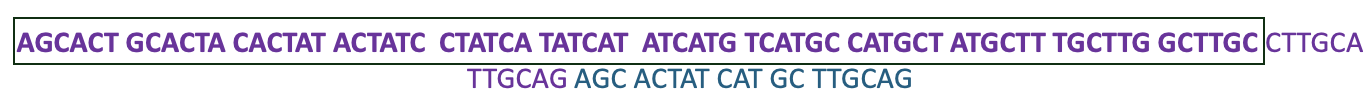}
\centering
\caption{Inconsistent token selection using traditional approach} 
\label{seq_sel}
\end{figure*}
\subsection{Data Preprocessing for Foundational Model Training}
To ensure uniformity and consistency in sequence processing for hybrid tokenization, each DNA sequence is first divided into fixed-length segments of 305 nucleotides before applying tokenization. This pre-segmentation step addresses the variability in token counts that may arise after tokenization, particularly for methods like Byte Pair Encoding (BPE), which do not produce a fixed number of tokens. Fig \ref{seq_sel} shows that in the previous method, selecting a fixed number of tokens (e.g., 12) could result in only selecting the k-mer part of the sequence, leaving out other essential tokens. To address this, we predetermine the sequence length to calculate the token length in a way that ensures all tokens for a given sequence are consistently included. For the 6-mer tokenization, the segmentation results in a deterministic number of tokens per segment, calculated as \(n - k + 1\) for a segment length of \(n = 305\) nucleotides and a k-mer size of \(k = 6\). To enhance model learning during Masked Language Modeling (MLM), we mask tokens with a probability of 0.15. When a token is selected for masking (denoted as position 0), the masking is extended to its surrounding positions \(-2, -1, 0, 1, 2, 3\), capturing the overlapping nature of k-mers. This ensures that the model focuses on learning robust representations from the context surrounding the chosen token, given the inherent overlaps in 6-mers. 
For the BPE tokenization, which produces a variable number of tokens per segment due to its flexible subword encoding approach, masking is applied independently with a probability of 0.15 to randomly selected tokens. This ensures consistency in the MLM objective across both tokenization schemes. By segmenting sequences prior to tokenization, maintaining fixed-length k-mer tokenization, and applying context-aware masking for overlapping k-mers, this hybrid approach combines the advantages of deterministic k-mer representation with the adaptiveness of BPE, facilitating effective learning of both local and global sequence patterns.

\subsection{Model Validation using Next k-mer Prediction}
To evaluate the efficiency of our proposed hybrid tokenization approach, we adopted the evaluation strategy introduced by Sanabria et al \cite{Sanabria2023} and further refined in the GROVER framework \cite{Sanabria2024}, which employs a fine-tuning task to identify the most effective foundational model. This method was specifically adapted to optimize the combination of k-mer and BPE tokenization methods for training the foundation model. In this evaluation approach, we first extracted the first 56 nucleotides from each given sequence, aligning with the methodology employed in GROVER. However, instead of tokenizing the sequence solely through one method, we applied our hybrid tokenization approach as described in Section \ref{hybrid-tokenization}, which combines both k-mer and BPE tokens. The remaining six nucleotides in the sequence were subsequently used to generate the next k-mer class label. For example, when predicting a 2-mer, the nucleotides at positions 51 and 52 serve as the class label, and the objective is to predict the next k-mer label from a pool of \(4k\) possible labels. In the tokenization process, we independently tokenize the input sequence with both k-mer and BPE methods and then concatenate the resulting tokens to form a unified tokenized representation. During training, we maintained a fixed sequence length of 80 tokens. This choice of length was based on the characteristics of the tokenization methods: for 6-mer tokenization (\(k=6\)), representing a sequence of 50 nucleotides requires no more than 45 tokens due to the overlapping structure of k-mers. Similarly, our experimental observations indicate that BPE tokenization does not exceed 30 tokens for sequences of this length. By setting a token limit of 80, we provide a sufficient buffer to accommodate both tokenization methods without truncating meaningful information, while ensuring a consistent input length across all sequences. This setup allows us to effectively evaluate and compare the performance of different tokenization combinations while preserving the integrity of the input data for the foundational model training.

\section{Experiments}
\label{experiments}
\subsection{Experimental Setup}
We conducted our experiments using an NVIDIA A100 80GB PCIe GPU. The training setup utilized the configuration (for both foundation and next-k-mer prediciton model training)  with a learning rate of 4e-4, Adam optimizer parameters set as epsilon=1e-6, beta1=0.9, and beta2=0.98, along with a weight decay of 0.01 to prevent overfitting. The model was trained for a maximum of 20,000 epochs with a gradient accumulation strategy using 25 steps, allowing efficient training with a batch size of 16 per device for training and 32 for evaluation. Checkpoints were saved every 2,500 steps, retaining up to 20 in total, and the best model was loaded at the end of training. Additionally, 1,000 warmup steps were included to stabilize training, while evaluation and logging occurred at intervals of 2,500 and 500 steps, respectively. 

\subsection{Dataset}
We used the Homo sapiens (human) genome assembly GRCh37 (hg19) \cite{GRCh37_hg19}. We only considered the sequences containing the nucleotides A, C, G, and T. We used 400k sequences for training and 100k sequences for testing.

\subsection{Foundation Model training Result}
\begin{figure}[h]
\includegraphics[width=0.5\textwidth]{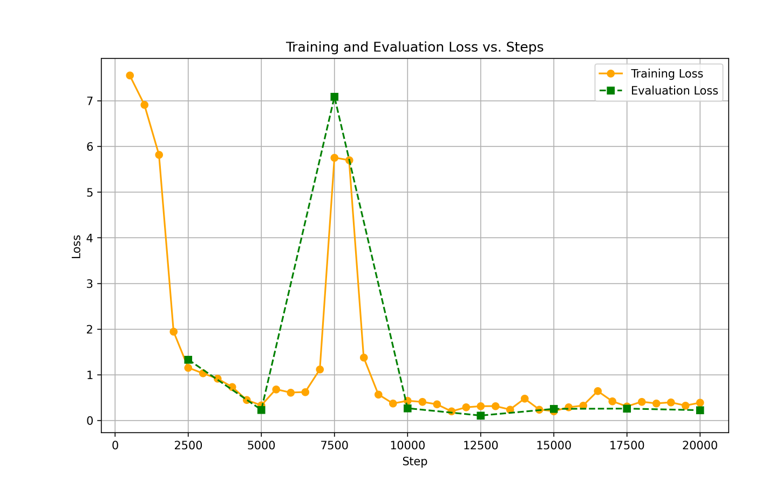}
\centering
\caption{The graph illustrates the trends in training and evaluation loss across training steps, showing how the loss values fluctuated and eventually stabilized.} 
\label{foundation_training}
\end{figure}

Fig. \ref{foundation_training} shows foundational model training results, highlighting training and evaluation loss metrics. The best evaluation loss achieved was 0.1072 at step 12,500 (epoch 1.8436), while the last was 0.2241 at step 20,000 (epoch 2.9497). The best training loss recorded was 0.1990 at step 11,500 (epoch 1.6961), with the last training loss being 0.3883 at step 20,000 (epoch 2.9497). 

\subsection{Fine Tuning Next-kmer Prediction}
We used the pre-trained Foundation DNA model and fine-tune it to predict the next k-mer, where k is 3, 4, 5.
\subsubsection{Data Preparation}
\begin{figure*}[!ht]
\includegraphics[width=1\textwidth]{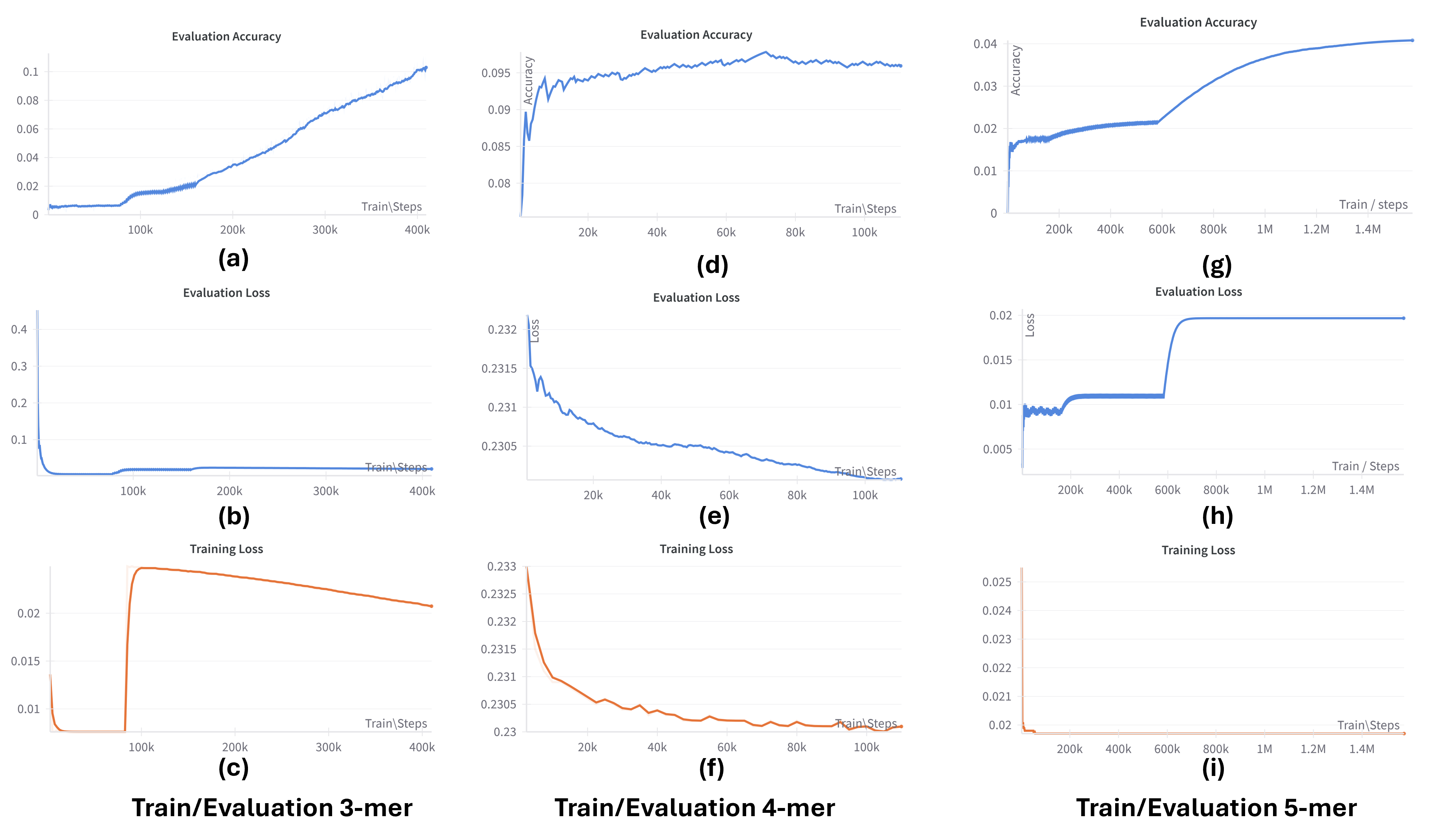}
\centering
\caption{Showing training and evaluation performance of Next 3,4 and 5-mer Prediction. Figs. (a), (b), and (c) represent the Evaluation accuracy, evaluation loss, and training loss of the 3-mer prediction task, respectively. Figs. (d), (e), and (f) represent the evaluation accuracy, evaluation loss, and training loss of the 4-mer prediction task, respectively. Figs. (g), (h), and (i) represent the evaluation accuracy, evaluation loss, and training loss of the 5-mer prediction task, respectively.} 
\label{train_val_results}
\end{figure*}

We conducted an experiment on the Ch21 dataset by dividing the genome into overlapping sequences of 510 nucleotides. From each 510-nucleotide sequence, only the first 56 nucleotides were retained. A total of 500,000 sequences were randomly selected, with 80\% allocated for training and 20\% for testing.  
For each sample, we used the first 50 nucleotides as input, while the labels corresponded to $4^k$ classes. Here, $k$ represents the number of nucleotides that follow the 50 input nucleotides, with each class representing a unique permutation of these subsequent nucleotides. This setup aimed to capture the dependencies and patterns within genomic sequences by predicting the combinations of nucleotide patterns that follow the input subsequences.
Training and validation performance of 3-mer, 4-mer and 5-mer prediction models are shown in fig. \ref{train_val_results} in terms of Evaluation accuracy, evaluation loss and training loss.

\subsection{Comparision with DLMs}
\begin{figure}[!ht]
\includegraphics[width=0.5\textwidth]{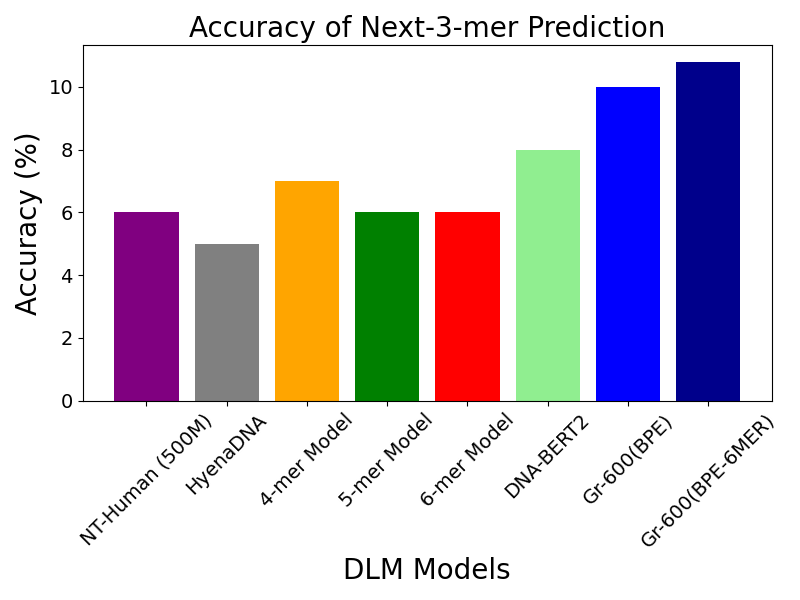}
\centering
\caption{A comparision of Next 3-mer Prediction accuracy of our model with NT-Human(500H), HyenaDNA, kmer models, DNA-BERT2 Grover(BPE-600)} 
\label{accuracy_3mer}
\end{figure}

Fig. \ref{accuracy_3mer} shows the comparison of 3-mer prediction accuracy across various deep learning models highlighting significant differences in performance, ranging from 5\% for HyenaDNA to 10.78\% for Gr-600 with BPE-6-MER. The proposed model based on BPE-600 and 6-mer tokenization (10.78\%) outperform others, showing that combining BPE-600 and 6-mer tokenization enhances accuracy. DNA-BERT2 achieves 8\%, reflecting its strong adaptation to DNA sequence tasks, while k-mer-based models (4-mer, 5-mer, and 6-mer) perform moderately, with the 4-mer model slightly leading at 7\%. In contrast, NT-Human and HyenaDNA lag behind with 6\% and 5\%, respectively, suggesting these models are less suited for fine-grained tasks like 3-mer prediction.

\begin{figure}[h]
\includegraphics[width=0.5\textwidth]{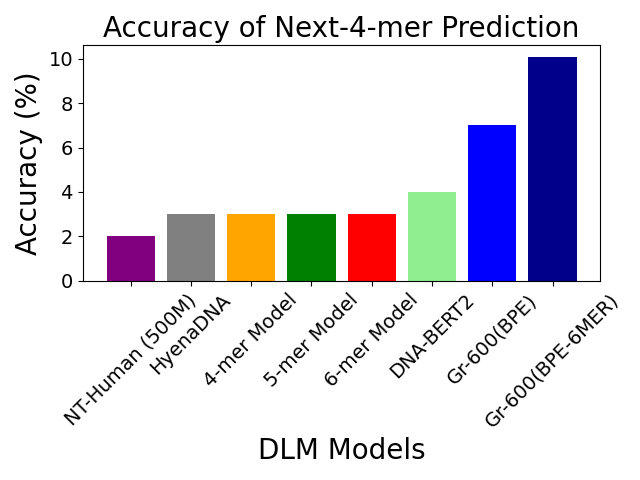}
\centering
\caption{A comparison of Next 4-mer Prediction accuracy of our model with NT-Human(500H), HyenaDNA, kmer models, DNA-BERT2 Grover(BPE-600)} 
\label{accuracy_4mer}
\end{figure}
The accuracy comparison of 4-mer prediction in Fig. \ref{accuracy_4mer} also shows that Gr-600 with BPE-6-MER leads with 10.1\%, followed by Gr-600 (BPE) at 7\%. DNA-BERT2 performs moderately well at 4\%, while other models, including HyenaDNA and kk-mer models (4-mer, 5-mer, and 6-mer), achieve around 3\%. NT-Human lags behind at 2\%.

\begin{figure}[h]
\includegraphics[width=0.5\textwidth]{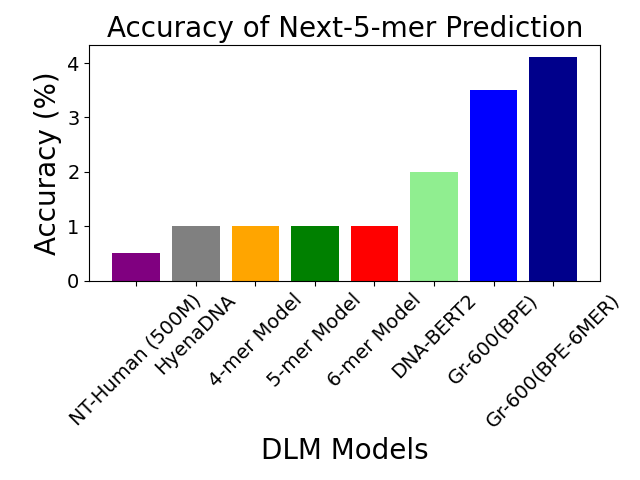}
\centering
\caption{A comparison of Next 5-mer Prediction accuracy of our model with NT-Human(500H), HyenaDNA, kmer models, DNA-BERT2 Grover(BPE-600)} 
\label{bpe_kmer_merging}
\label{accuracy_5mer}
\end{figure}

The comparison of 5-mer prediction accuracy in fig. \ref{accuracy_5mer} shows that Gr-600 with BPE-6-MER achieves the highest accuracy at 4.12\%, followed by Gr-600 (BPE) at 3.5\% and DNA-BERT2 at 2\%. Other models, including HyenaDNA and k-mer models (4-mer, 5-mer, and 6-mer), perform similarly at 1\%, while NT-Human trails at 0.5\%. 

These results emphasize the importance of tokenization strategies and architecture in improving prediction accuracy, with hybrid methods like Gr-600 with BPE-6-MER offering the best performance.

\section{Limitation and Challanges}
\label{limitations}
We faced several challenges in training the foundation model, primarily due to the vast complexity and scale of genomic data. High computational costs and time requirements made it difficult to access the necessary GPU resources for training. Additionally, DNA sequences are often too long for standard models, resulting in a loss of context when split into smaller fragments while it is needed to concatenate k-mer and BPE for the same sequence leads to larger sequences, further complicating the training process.

\section{Conclusion}
\label{conclusion}
This research essentially dives into the problems with traditional tokenization methods, looks at the strengths and weaknesses of models like NT, DNABERT, GROVER, and k-mer-based models, and highlights the experimental findings of applying the proposed hybrid tokenization (BPE+K-MER) approach to train the foundation model. We evaluated the performance of our foundation model trained using the optimal vocabulary obtained using BPE-600 and 6-mer tokenization through the accuracy of next-k-mer prediction as a fine-tuning task for 3, 4, and 5-mer. The experimental results show the best performance in next-k-mer prediction while applying the hybrid tokenization strategy in training the foundation DLM model.

\section{Future Works}
\label{future_works}
We plan to extend the training and evaluation to include 2-mer and 6-mer tokenizations in future work. Subsequently, we will explore additional combinations of k-mers (2 and 6) along with BPE-600 to identify the optimal vocabulary configuration for foundation model training. Our foundation model will be rigorously assessed across multiple downstream tasks, including promoter identification (Prom-300), promoter scanning (Prom Scan), transcription factor binding site prediction (CTCF motif binding), and splice site prediction, to evaluate its performance on diverse genomic datasets.
Moreover, we will conduct fine-tuning experiments on state-of-the-art DNA language models (DLMs) such as DNABERT2, NT, HyenaDNA, and other k-mer-based models. Comparative analyses will be performed to benchmark the performance of our model against these existing approaches on the specified downstream tasks. 
Given the increased sequence length resulting from the integration of two tokenization techniques, we will also investigate training tokenization strategies on DLMs capable of handling long sequences. This will address the challenges posed by larger token sequences, ensuring efficient training and evaluation while maintaining model performance across complex genomic datasets.

%%%%%%%%%%%%%%%%%%include references%%%%%%%%%%%%%%%%%%%%%%%%%%%%%%%%%%
{
    \small
    % \bibliographystyle{ieeenat_fullname}
    % \bibliography{main}

    \bibliographystyle{ieeetr}
    \bibliography{references.bib}

\begin{thebibliography}{10}

\bibitem{Nguyen2023}
E.~Nguyen, M.~Poli, M.~Faizi, A.~Thomas, C.~Birch-Sykes, M.~Wornow, A.~Patel, C.~Rabideau, S.~Massaroli, Y.~Bengio, S.~Ermon, S.~A. Baccus, and C.~Ré, ``Hyenadna: Long-range genomic sequence modeling at single nucleotide resolution,'' {\em ArXiv [Preprint]}, November 14 2023.
\newblock Preprint.

\bibitem{Avsec2021}
Å.~Avsec, V.~Agarwal, D.~Visentin, J.~R. Ledsam, A.~Grabska-Barwinska, K.~R. Taylor, Y.~Assael, J.~Jumper, P.~Kohli, and D.~R. Kelley, ``Effective gene expression prediction from sequence by integrating long-range interactions,'' {\em Nature Methods}, vol.~18, pp.~1196--1203, October 2021.

\bibitem{10.1093/bioinformatics/btab083}
Y.~Ji, Z.~Zhou, H.~Liu, and R.~V. Davuluri, ``Dnabert: pre-trained bidirectional encoder representations from transformers model for dna-language in genome,'' {\em Bioinformatics}, vol.~37, pp.~2112--2120, 02 2021.

\bibitem{NT_Dalla-Torre2023.01.11.523679}
H.~Dalla-Torre, L.~Gonzalez, J.~Mendoza-Revilla, N.~Lopez~Carranza, A.~Henryk~Grzywaczewski, F.~Oteri, C.~Dallago, E.~Trop, B.~P. de~Almeida, H.~Sirelkhatim, G.~Richard, M.~Skwark, K.~Beguir, M.~Lopez, and T.~Pierrot, ``The nucleotide transformer: Building and evaluating robust foundation models for human genomics,'' {\em bioRxiv}, 2024.

\bibitem{LOGO_Yang2022}
M.~Yang, L.~Huang, H.~Huang, H.~Tang, N.~Zhang, H.~Yang, J.~Wu, and F.~Mu, ``Integrating convolution and self-attention improves language model of human genome for interpreting non-coding regions at base-resolution,'' {\em Nucleic Acids Research}, vol.~50, p.~e81, August 2022.
\newblock © The Author(s) 2022. Published by Oxford University Press on behalf of Nucleic Acids Research.

\bibitem{devlin2019bertpretrainingdeepbidirectional}
J.~Devlin, M.-W. Chang, K.~Lee, and K.~Toutanova, ``Bert: Pre-training of deep bidirectional transformers for language understanding,'' 2019.

\bibitem{DNAGPT}
D.~Zhang, W.~Zhang, B.~He, J.~Zhang, C.~Qin, and J.~Yao, ``Dnagpt: A generalized pretrained tool for multiple dna sequence analysis tasks,'' {\em bioRxiv}, 2023.

\bibitem{schiff2024caduceusbidirectionalequivariantlongrange}
Y.~Schiff, C.-H. Kao, A.~Gokaslan, T.~Dao, A.~Gu, and V.~Kuleshov, ``Caduceus: Bi-directional equivariant long-range dna sequence modeling,'' 2024.

\bibitem{pióro2024moemambaefficientselectivestate}
M.~Pióro, K.~Ciebiera, K.~Król, J.~Ludziejewski, M.~Krutul, J.~Krajewski, S.~Antoniak, P.~Miłoś, M.~Cygan, and S.~Jaszczur, ``Moe-mamba: Efficient selective state space models with mixture of experts,'' 2024.

\bibitem{li2024vqdnaunleashingpowervector}
S.~Li, Z.~Wang, Z.~Liu, D.~Wu, C.~Tan, J.~Zheng, Y.~Huang, and S.~Z. Li, ``Vqdna: Unleashing the power of vector quantization for multi-species genomic sequence modeling,'' 2024.

\bibitem{zhou2024dnabert2efficientfoundationmodel}
Z.~Zhou, Y.~Ji, W.~Li, P.~Dutta, R.~Davuluri, and H.~Liu, ``Dnabert-2: Efficient foundation model and benchmark for multi-species genome,'' 2024.

\bibitem{bostrom2020bytepairencodingsuboptimal}
K.~Bostrom and G.~Durrett, ``Byte pair encoding is suboptimal for language model pretraining,'' 2020.

\bibitem{Sanabria2024}
M.~Sanabria, J.~Hirsch, P.~M. Joubert, and A.~R. Poetsch, ``Dna language model grover learns sequence context in the human genome,'' {\em Nature Machine Intelligence}, vol.~6, pp.~911--923, August 2024.

\bibitem{Sanabria2023}
M.~Sanabria, J.~Hirsch, and A.~R. Poetsch, ``Distinguishing word identity and sequence context in dna language models,'' {\em bioRxiv}, 2023.

\bibitem{GRCh37_hg19}
N.~C. for Biotechnology Information~(NCBI), ``Homo sapiens genome assembly grch37 (hg19),'' 2009.
\newblock Accessed: 2024-12-13.

\end{thebibliography}
}
\end{document}